\pdfoutput=1

\documentclass[11pt]{article}

\usepackage[]{emnlp2021}

\usepackage{times}
\usepackage{latexsym}
\usepackage{graphicx}
\usepackage{comment}
\usepackage{amsmath,amssymb,bm} 
\usepackage{color}
\usepackage{xcolor}
\usepackage{lipsum}
\usepackage{tabularx}
\usepackage{adjustbox}
\usepackage{multirow}
\usepackage{siunitx}
\usepackage{dsfont}
\usepackage{xcolor}
\usepackage{array, booktabs, bigdelim, makecell}
\usepackage{floatrow}
\usepackage{url}
\newfloatcommand{capbtabbox}{table}[][\FBwidth]

\newcommand{\fig}{Figure}
\newcommand{\tbl}{Table}
\newcommand{\eg}{\textit{e.g.}}
\newcommand{\ie}{\textit{i.e.}}
\newcommand{\sect}{Sec.}
\newcommand\blfootnote[1]{%
  \begingroup
  \renewcommand\thefootnote{}\footnote{#1}%
  \addtocounter{footnote}{-1}%
  \endgroup
}

\usepackage[T1]{fontenc}

\usepackage[utf8]{inputenc}

\usepackage{microtype}
\usepackage{parskip}
\title{Reasoning Visual Dialog with Sparse Graph Learning and \\ Knowledge Transfer}

\author{Gi-Cheon Kang \\ Seoul National University, AIIS \\ \normalsize{\texttt{chonkang@snu.ac.kr}} \And
         Junseok Park \\ Seoul National University \\ \normalsize{\texttt{jspark227@snu.ac.kr}} \And
         Hwaran Lee \\ NAVER AI Lab \\ \normalsize{\texttt{hwaran.lee@navercorp.com}} \\ 
         \AND
         Byoung-Tak Zhang$^\dagger$ \\ Seoul National University, AIIS \\ \normalsize{\texttt{btzhang@snu.ac.kr}} \And
         Jin-Hwa Kim$^\dagger$ \\ NAVER AI Lab \\ \normalsize{\texttt{j1nhwa.kim@navercorp.com}} \\} 

\begin{document}
\maketitle
\begin{abstract}
Visual dialog is a task of answering a sequence of questions grounded in an image using the previous dialog history as context. In this paper, we study how to address two fundamental challenges for this task: (1) reasoning over underlying semantic structures among dialog rounds and (2) identifying several appropriate answers to the given question. To address these challenges, we propose a Sparse Graph Learning (SGL) method to formulate visual dialog as a graph structure learning task. SGL infers inherently sparse dialog structures by incorporating binary and score edges and leveraging a new structural loss function. Next, we introduce a Knowledge Transfer (KT) method that extracts the answer predictions from the teacher model and uses them as pseudo labels. We propose KT to remedy the shortcomings of single ground-truth labels, which severely limit the ability of a model to obtain multiple reasonable answers. As a result, our proposed model significantly improves reasoning capability compared to baseline methods and outperforms the state-of-the-art approaches on the VisDial v1.0 dataset. The source code is available at \url{https://github.com/gicheonkang/SGLKT-VisDial}.
\end{abstract}

\section{Introduction}
\blfootnote{$\dagger$ corresponding authors.}
Recently, visually-grounded dialogue \cite{das2017visual,de2017guesswhat,Kottur2019CLEVRDialog,kim2017codraw} has attracted increasing research interest due to its potential impact on many real-world applications (\eg, aiding visually impaired user). Notably, Visual Dialog (VisDial) \cite{das2017visual}, which extends visual question answering (VQA) \cite{antol2015vqa,kim2018bilinear,seo2021attend} to multi-round dialog, has been introduced to the research community, along with a large scale dataset. Unlike VQA, VisDial is designed to answer a \emph{sequence} of questions grounded in an image utilizing a dialog history as context. This task requires a deep understanding of multi-modal inputs and the temporal nature of a human conversation. To infer an appropriate answer to the question, a dialog agent should attend to meaningful context from the dialog history as well as the given image.

There are two fundamental challenges in VisDial: (1) reasoning over underlying semantic structures among a series of utterances (\ie, dialog rounds) and (2) identifying several appropriate answers to the given question. Previous approaches have implicitly addressed the first challenge by using the \emph{soft-attention mechanism} \cite{bahdanau2014neural}. Typically, the soft-attention mechanism is utilized to discover semantic relationships between the given question and previous utterances (\ie, dialog history) while extracting rich contextual representations \cite{gan2019multi,agarwal2020history}. Next, most of the previous work has not explicitly tackled the second challenge since there are no labels for prediction of multiple possible answers. For this reason, they have mostly focused on finding the single ground-truth answer by leveraging standard one-hot encoded labels.  

\begin{figure*}[t!]
\centering
\includegraphics[width=0.98\textwidth]{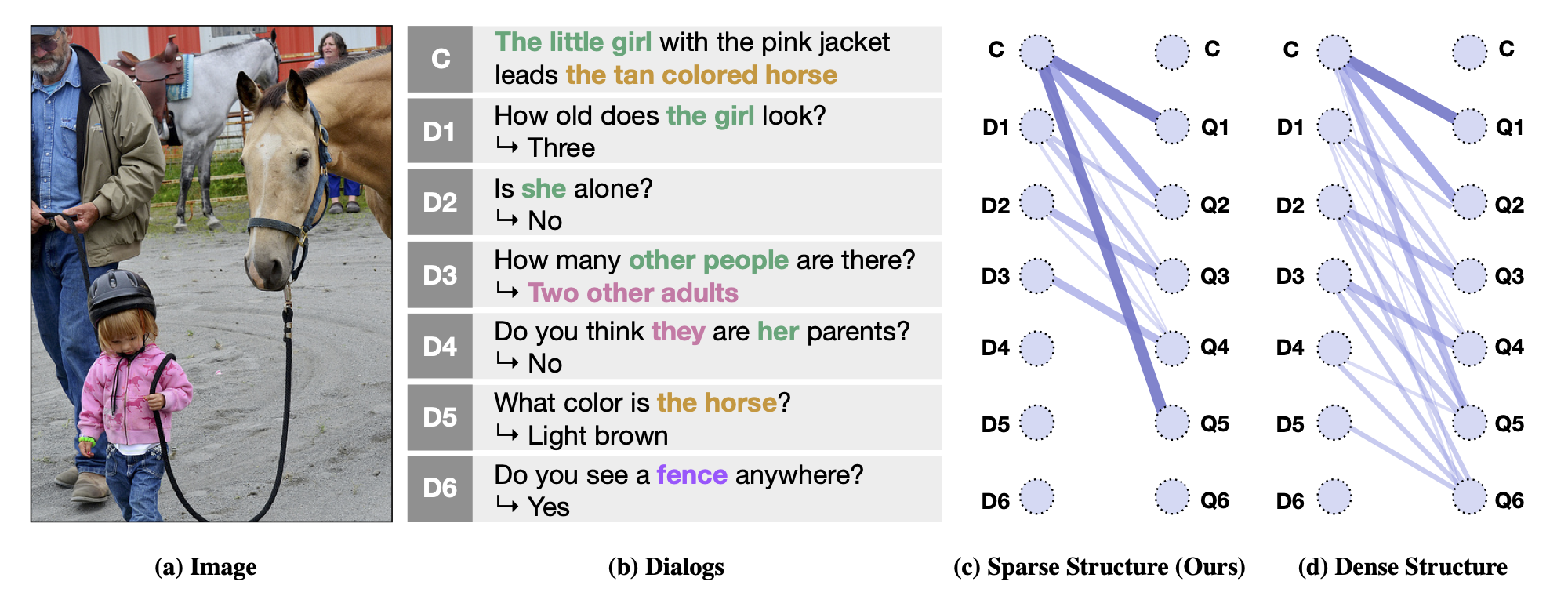}
\caption{An example from the VisDial dataset. (a): a given image. (b): dialogue regarding the image, including image caption (C), and each round of dialog (D1-D6). (c) and (d): the semantic structures from our proposed model and the soft attention-based model, respectively. The left and right column in each figure denote the dialog history and the current question, respectively. The thicker and darker links indicate the higher semantic dependencies.}
\label{fig:example}
\end{figure*}

We argue that existing approaches in VisDial show limited reasoning capability due to the way they approach the task: \emph{soft-attention} and \emph{one-hot encoded labels}. First, soft-attention restricts the ability to represent various types of semantic relationships in the dialog. As we illustrate in \fig~\ref{fig:example}, some questions in the dialog (Q1-Q4) are semantically dependent on previous utterances, while others (Q6) are independent, due to an abrupt change in topic. Furthermore, previous topics could be readdressed later in the dialog (Q5). However, soft-attention, which is based on the softmax function, always assigns a \textit{non-zero} weight to all previous utterances, which results in dense (\ie, fully-connected) relationships. Moreover, the sum of attention weights should be \textit{one} due to the sum-to-1 constraint of the softmax function. Herein lies the problem: even for questions that are partly dependent (Q5 in \fig~\ref{fig:example}) or independent (Q6 in \fig~\ref{fig:example}) from the dialog history, all previous utterances are still considered and integrated into the contextual representations. As a consequence, the dialog agent could overly rely on the dialog history, even when the dialog history is irrelevant to the given question. Second, the model that utilizes the one-hot encoded labels learns to predict the single ground-truth answer only. However, similar to VQA, the given question is associated with one or several answers from a set of candidate answers. Therefore, the one-hot labels could suppress several plausible answers, assigning unreasonably low prediction probabilities to them.  

In this paper, we propose two methods to remedy the conceptual shortcomings of the current approaches discussed above. First, we introduce a Sparse Graph Learning (SGL) method that predicts sparse structures of the visually-grounded dialog. In the graph structure, each node corresponds to a round of the dialog, and edges represent the semantic relationships between the rounds. SGL constructs the representations of each node by embedding the given image and each round of dialog in a joint fashion. SGL then infers two types of edge weights: binary (\ie, 0 or 1) and score edges. It ultimately discovers the sparse and weighted structures (\eg, (c) in \fig~\ref{fig:example}) by incorporating the two edge weights. Furthermore, we design a new structural loss function to encourage SGL to infer explicit and reliable dialog structures by leveraging a structural supervision. Next, to identify multiple possible answers, we treat VisDial as a regression task that predicts the correctness of each candidate answer individually, instead of a traditional setting that estimates the sum-to-1 scores over the candidate answers. To this end, we propose a Knowledge Transfer (KT) method that extracts the soft scores of each candidate answer from the teacher model \cite{qi2020two}. The soft scores are used to optimize for multiple possible answers. We expect this work to shed light on the above challenges that have not been explicitly addressed in visual dialog. 

The main contributions of our paper are as follows. First, we propose a Sparse Graph Learning (SGL) approach that builds sparse structures of the visually-grounded dialog. By leveraging a new structural loss function, SGL learns the semantic relationships among dialog rounds in an explicit way. Second, we introduce a Knowledge Transfer (KT) method to encourage the model to find multiple possible answers to the given question. Third, the model that utilizes SGL and KT achieves the new state-of-the-art results on the VisDial v1.0 dataset. We perform comprehensive analysis to validate the effectiveness of SGL and KT. Finally, we conduct a qualitative analysis of each proposed method.  

\section{Related Work}
\noindent\textbf{Visual Dialog} \cite{das2017visual} has been introduced as a temporal extension of VQA \cite{antol2015vqa}. In this task, a dialog agent should answer a sequence of questions by using an image and the dialog history as a clue. We carefully categorize the previous studies on visual dialog into three groups: (1) soft attention-based methods that compute the interactions among entities, including an input image, questions, and dialog history \cite{gan2019multi,schwartz2019factor,agarwal2020history,murahari2019large,wang2020vd}, (2) a visual coreference resolution method \cite{seo2017visual,kottur2018visual,niu2018recursive,kang2019dual} that clarifies ambiguous expressions (\eg, it, them) in the question and links them to the specific entities in the image, and (3) a structural inference method \cite{zheng2019reasoning} that attempts to discover dialog structures based on graph neural networks. Our approach belongs to the third group. Similar to the soft attention-based methods, \citet{zheng2019reasoning} infer the dense semantic structures using a softmax function. Moreover, they attempt to find the structures without any explicit optimization for the structural inference. To tackle these aspects, we propose SGL which explicitly infers sparse structures with a structural loss function.

\noindent\textbf{Graph Neural Networks} \cite{scarselli2008graph} have sparked a tremendous interest at the intersection of deep neural networks and structural learning approaches. Recently, graph learning networks (GLNs) were proposed by \cite{pilco2019graph,on2020cut}, with the goal of reasoning over underlying structures of input data. GLNs consider unstructured data and dynamic domains (\textit{e.g}., time-varying domain). Our method belongs to the group of GLNs. CB-GLNs \cite{on2020cut} attempt to discover the compositional structure of long video data with a graph-cut algorithm \cite{shi2000normalized}. However, SGL is different from previous studies in that SGL learns to build sparse structures \textit{adaptively}, not relying on a predefined algorithm, and the dataset we use is highly multimodal. 

\noindent\textbf{Knowledge Transfer} technique has been mainly explored to compress a large model into a small model \cite{bucilua2006model,ba2014deep} without a significant drop in accuracy. The idea of knowledge transfer was later popularized under the name of knowledge distillation (KD) \cite{Hinton2014}. In KD, the knowledge of the large model (\ie, teacher model) is transferred to the small model (\ie, student model) as a form of supervision signal. Then, the student model learns to mimic the behavior of the teacher model by using the supervision signal and a pre-defined distillation loss function. Our Knowledge Transfer (KT) approach shares this same spirit. However, we repurpose KT to cast VisDial as a regression of scores for candidate answers. Accordingly, the soft targets from the teacher model are utilized as supervision for the correctness of each candidate answer which was originally unlabeled.

\section{Sparse Graph Learning}
The visual dialog task \cite{das2017visual} is defined as follows: given an image $\mathcal{I}$, a caption $c$ describing the image, a dialog history $\mathcal{H} = \{\underbrace{c}_{h_0}, \underbrace{(q_1, a^{gt}_1)}_{h_1}, \cdots , \underbrace{(q_{t-1}, a^{gt}_{t-1})}_{h_{t-1}}\}$, and a question $q_t$ at current round $t$, the goal is to find an appropriate answer to the question among the $N$ answer candidates, {\it $\mathcal{A}_t$ = $\left\{a_{t}^1, \cdots, a_{t}^{N} \right\}$}. 

In our approach, we consider the task as a graph $G_t = (V_t, E_t)$ with $t+1$ nodes (\ie, vertices), where $(v_0, v_1,...,v_{t-1})$ and $(v_t)$ correspond to the node for the previous dialog history and the current question, respectively. Each node $v_i \in V_t$ is associated with a feature vector $\mathbf{x}_i$. The semantic dependencies among the nodes are represented as weighted edges $E_t = \left\{(v_i, v_j): v_i, v_j \in V_t\right\}$. The goal of our approach is to discover a sparse and weighted adjacency matrix $\mathbf{A}_t \in \mathbb{R}^{(t+1) \times (t+1)}$ which represents the semantic dependencies among dialog rounds. 

\begin{figure*}[h!]
\centering
\includegraphics[width=0.98\textwidth]{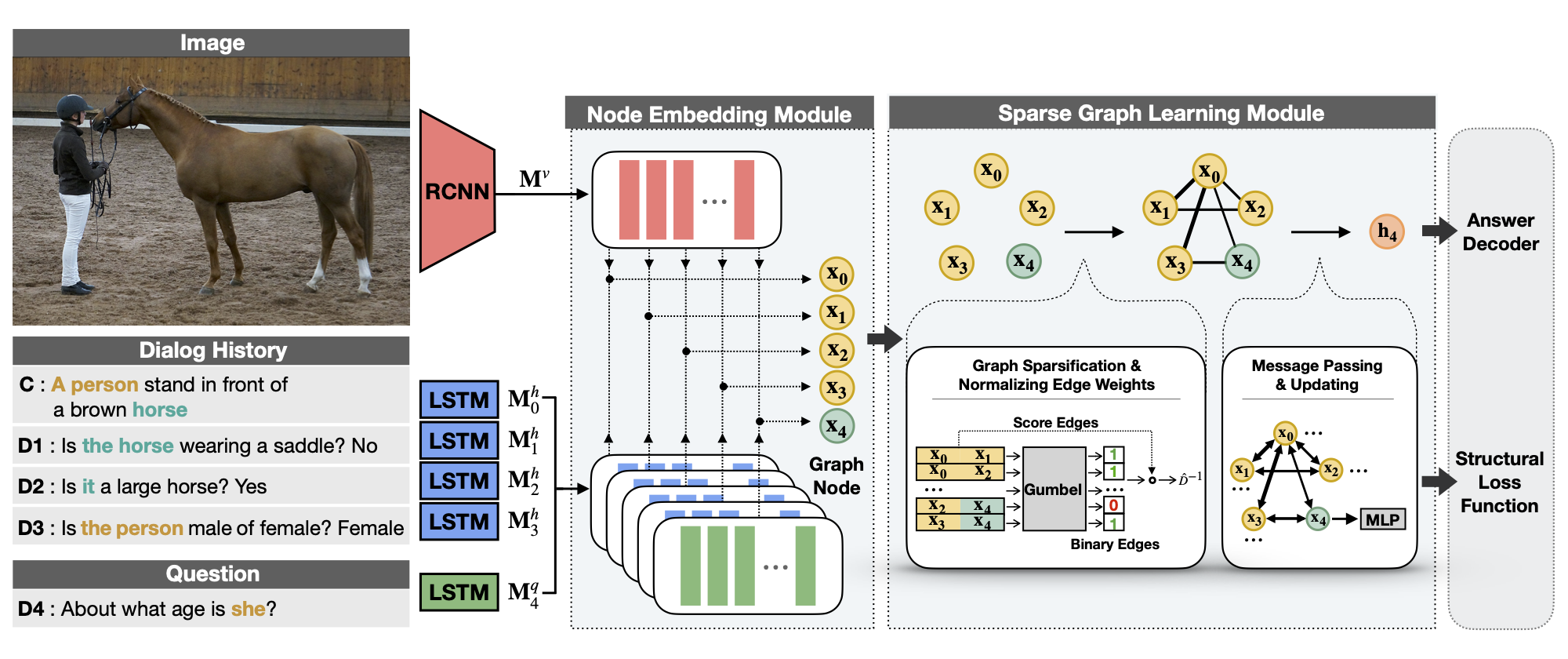}
\caption{An overview of Sparse Graph Learning (SGL) framework. Please see Section 3 for details.}
\label{fig:overview}
\end{figure*}

To implement the pipeline above, we propose a Sparse Graph Learning (SGL) method that consists of two modules (see \fig~\ref{fig:overview}): (1) a node embedding module that embeds the visual-linguistic representations for each round of the dialog and (2) a sparse graph learning module that estimates a sparse and weighted structures of the dialog. 

\subsection{Input Features}
\label{sec:input}
\noindent\textbf{Visual Features.} In the given image $\mathcal{I}$, we extract the $d_v$-dimensional visual features of $K$ objects by employing a pre-trained Faster R-CNN model \cite{ren2015faster,Anderson2017up-down}. Then, we project the visual features into dimension $d_h$ using a linear matrix $\mathbf{W}_f \in \mathbb{R}^{d_v \times d_h}$, which results in $ \mathbf{M}^{v} \in$ $\mathbb{R}^{K \times d_h}$. We use $ \mathbf{M}^{v}$ as visual features. \\

\noindent\textbf{Language Features.} In the $t$-th dialog round, we first encode the question $q_t$ which is a word sequence of length $L$, $({w}_{1}, ..., {w}_{L})$, by using a LSTM \cite{hochreiter1997long}. Specifically, we use all hidden states of the LSTM as the question features, which results in $\mathbf{M}^{q}_t \in \mathbb{R}^{L \times d_h}$. Likewise, each round of the dialog history $\left\{h_i \right\}_{i=0}^{t-1}$ is encoded into $\left\{\mathbf{M}^{h}_i \right\}_{i=0}^{t-1}\in \mathbb{R}^{t \times L \times d_h}$. To reduce computational complexity, we embed all the answer candidates $\left\{a_t^i \right\}_{i=1}^{N}$ with sentence-level features by extracting the last hidden states of the LSTM, which results in $\mathbf{M}_{t}^{a} \in \mathbb{R}^{N \times d_h}$.

\subsection{Node Embedding Module}
The node embedding module aims to embed rich visual-linguistic joint representations for each round of the dialog. To implement these processes, we take inspiration from Modular Co-Attention Networks (MCAN) \cite{yu2019deep} which are based on the multi-head attention mechanism \cite{vaswani2017attention}. Given the object-level visual features $ \mathbf{M}^v \in$ $\mathbb{R}^{K \times d_h}$ and the question features $\mathbf{M}^q_t \in$ $\mathbb{R}^{L \times d_h}$, the node embedding module $f_{ne}$ computes the joint representations $\mathbf{x}_t \in \mathbb{R}^{1 \times d_h}$.
\begin{align}
    \mathbf{x}_t = f_{ne} (\mathbf{M}^v, \mathbf{M}^q_t) 
\end{align}
Each round of the dialog history $\left\{\mathbf{M}^{h}_i \right\}_{i=0}^{t-1}$ is also embedded by the module, which results in $\left\{\mathbf{x}_i \right\}_{i=0}^{t-1}$. Consequently, as shown in Figure~\ref{fig:overview}, we obtain ($t+1$) joint representations including the question features $\mathbf{x}_t$ and the dialog features $\left\{\mathbf{x}_i \right\}_{i=0}^{t-1}$. We use these features as the nodes of the graph which can be represented in matrix-form as $\mathbf{X} \in \mathbb{R}^{(t+1) \times d_h}$. A detailed architecture of the node embedding module can be found in the supplementary materials.


\subsection{Sparse Graph Learning Module}
The sparse graph learning module infers the underlying sparse and weighted graph structure among nodes, where the edge weights are estimated based on the node features. To make the graph structure to be sparse, we propose two types of edges on the graph $G_t$: binary edges $E_t^b$ and score edges $E_t^s$, whose corresponding adjacency matrices are $\mathbf{A}_t^b$ and $\mathbf{A}_t^s$ respectively. To simplify the notation, we omit the subscript $t$ in the following equations. \\

\noindent\textbf{Binary Edges.}
We first define a binary edge between two nodes $v_i$ and $v_j$ as a binary random variable $z_{ij}\in \left\{ 0,1 \right\}$, for all $ i,j\in [0, t]$ and $i < j$. The sparse graph learning module estimates the likelihood of the binary variables given the node features, where the probability implies whether the two nodes are semantically related or not. We regard the binary variable as a two-class categorical variable and define the probability distribution as: 
\begin{align}
    \mathbf{A}_{ij}^b &= z_{ij} \sim Categorical(\mathbf{p}_{ij}) \\
    \mathbf{p}_{ij} &= \mathrm{softmax}\Big(\mathbf{W}_c (\mathbf{x}_i \circ \mathbf{x}_j)^\top / \tau \Big)
\end{align}
where $ \mathbf{W}_c \in \mathbb{R}^{2 \times d_h}$ is a learnable parameter, $\circ$ denotes the hadamard product, and $\tau$ is the softmax temperature.
Since $z_{ij}$ is discrete and non-differentiable, we employ a Straight-Through Gumbel-Softmax estimator (\textit{i.e}., ST-Gumbel) \cite{jang2016categorical} to ensure end-to-end training.
During forward propagation, ST-Gumbel makes a discrete decision by using the Gumbel-Max trick:
\begin{align} \label{eq:gumbel-max}
z_{ij} &=
    \begin{cases}
        1, & \text{if } \underset{k \in \left\{0,1\right\} } {\mathrm{argmax}}\big(\mathrm{log}(p_k) + g_k\big) = 1 \\
        0, & \text{otherwise}
    \end{cases}
\end{align}
where the random variable $g_k$ is drawn from a Gumbel distribution. In the backward pass, ST-Gumbel utilizes the derivative of the probabilities by approximating $\nabla_{\theta}z \approx \nabla_{\theta}p$, thus enabling the backpropagation and end-to-end training. \\

\noindent\textbf{Score Edges.}
We define score edges to measure the extent to which two nodes are related, and the relevance is computed as:
\begin{align}
    \mathbf{A}_{ij}^s &= (\mathbf{x}_i \mathbf{x}_j^\top)^2
\end{align}
Following \citet{yang2018glomo}, we also employ the squared dot product for stabilized training. \\

\noindent\textbf{Sparse Weighted Edges.}
The sparse graph learning module multiplies the binary edges and score edges, finally yielding a sparse and weighted adjacency matrix as:
\begin{align}
    \hat{\mathbf{A}}_{ij} &= \mathbf{A}_{ij}^b \mathbf{A}_{ij}^s = z_{ij} (\mathbf{x}_i \mathbf{x}_j^\top)^2
\end{align}
With the above edge weight estimations, this module is able to model three types of relationships on $v_i$: (1) dense relationships similar to the previous conventional softmax-based approaches if $\sum_j{z_{ij}}=t$ (\textit{i.e}., all entries in $z_{i}$ are one), 
(2) sparse relationships if $0<\sum_j{z_{ij}}<t$, and 
(3) no relationships if $\sum_j{z_{ij}}=0$ (\textit{i.e., isolated node)}. \\

\noindent\textbf{Message-passing and Update.}
Based on the sparse weighted adjacency matrix $\hat{\mathbf{A}}$, the sparse graph learner updates the hidden states of all nodes through a message-passing framework \cite{gilmer2017neural}. Similar to graph convolutional networks \cite{kipf2016semi}, we simply implement the message-passing layer $F_M$ as the normalized weighted sum according to the adjacent weight, followed by a linear transformation. 
\begin{align}
\mathbf{M} &= F_M(\mathbf{X}, \hat{\mathbf{A}}) = \hat{\mathbf{D}}^{-1} \hat{\mathbf{A}} \mathbf{X} \mathbf{W}_m
\label{eq:message}
\end{align}
where $\mathbf{W}_m \in \mathbb{R}^{d_h \times d_h}$. Note that $\hat{\mathbf{D}}$ is the degree matrix of $\hat{\mathbf{A}}$. The hidden node features are calculated via the update layer $F_U$ which adds the input feature and aggregated messages and subsequently feeds them into a non-linear function $f_u$.
\begin{align}
\mathbf{H} &= F_U(\mathbf{X}, \mathbf{M}) = f_u (\mathbf{X} + \mathbf{M})
\label{eq:update}
\end{align}
$f_u$ is two-layer feed-forward networks with a ReLU in between. The model can perform multi-step reasoning by conducting a set of equations (\ie, Eq. \ref{eq:message} and Eq. \ref{eq:update}) multiple times. Finally, SGL returns the adjacency matrix $\hat{\mathbf{A}}$ and the hidden node features $\mathbf{H} \in \mathbb{R}^{(t+1) \times d_h}$. The features for the current round, $\mathbf{H}{[t,:]} = \mathbf{h}_t$, is used to decode answers. Note that SGL as described above computes all interactions among $t+1$ nodes for every dialog round, although the edge weights among $\left\{ \mathbf{x}_i \right\}_{i=0}^{t-1}$ are estimated in the previous dialog round. For the sake of computational efficiency, we can construct $\hat{\mathbf{A}}_t$ by combining the adjacency matrix of the previous round $\hat{\mathbf{A}}_{t-1}$ with the edge weights between $\mathbf{x}_t$ and $\left\{ \mathbf{x}_i \right\}_{i=0}^{t-1}$ in the $t$-th round. This decreases the computational complexity, from $O(t^2)$ to $O(t)$. 

\subsection{Structural Learning}
We introduce a structural loss function $\mathcal{L}_{sgl}$ to encourage SGL to infer explicit, reliable dialog structures. Inspired by Coref-NMN \cite{kottur2018visual} that employs the off-the-shelf neural coreference resolution tool\footnote{\href{https://github.com/huggingface/neuralcoref}{https://github.com/huggingface/neuralcoref} based on the work \cite{clark2016deep}.} for visual coreference resolution, we repurpose this tool for structural learning. Specifically, we automatically obtain the semantic dependencies between rounds by using the coreference resolution tool and leverage this information as structural supervision. The one-valued entries in the structural supervision indicate that both dialog rounds include at least one noun phrase or a pronoun referring to the same entity. Otherwise, the entries are filled with a zero-value. SGL minimizes the distance between the structural supervision $\mathbf{C}_T$ and the binary matrix $\mathbf{A}_T^b \in \mathbb{R}^{(T+1) \times (T+1)}$ finally predicted from SGL:
\begin{align}
\mathcal{L}_{sgl} &= {\parallel \mathbf{C}_T-\mathbf{A}^b_T\parallel}^2_F
\end{align}
where $T$ and ${\parallel \cdot \parallel}^2_F$ denote the total number of rounds for each dialog and the squared Frobenius norm (\ie, element-wise mean squared error), respectively. Here, $\mathcal{L}_{sgl}$ encourages SGL to predict a reliable dialog structure. Note that SGL uses the structural supervision only while training, and infers the dialog structures at test time.

\section{Knowledge Transfer}
\label{sec:knowledge_transfer}
The conventional assumption in VisDial is that there is one correct answer for each question from a set of candidate answers. Accordingly, the one-hot encoded single ground-truth label is used as standard supervision. However, the given question can indeed be associated with one or several answers. For this reason, a few works \cite{qi2020two,murahari2019large} have applied an additional fine-tuning strategy on dense labels\footnote{The densely annotated relevance scores for all candidate answers are released in the VisDial v1.0 validation \& test split.} for the validation split to improve the model's ability to predict multiple correct answers. Instead of using the fine-tuning approach, we propose a Knowledge Transfer (KT) method to optimize several correct answers simultaneously in a single training procedure. KT extracts the soft scores of each candidate answer from the fine-tuned teacher model, P1+P2 \cite{qi2020two}, and uses these scores as pseudo labels. We choose the P1+P2 for their strong performance on retrieving several appropriate answers for the given question. Specifically, we combine the dense score vector $\mathbf{y}_t^{dense} \in \mathbb{R}^{N}$ from the teacher model with the one-hot vector $\mathbf{y}_t^{sparse} \in \mathbb{R}^{N}$ for the $t$-th question as:
\begin{align}
    \hat{y}_{tn} = \underset{n \in \left\{1, ... , N\right\} } {\mathrm{max}}\big(y_{tn}^{sparse}, \; y_{tn}^{dense}\big) 
\end{align}
where $N$ is the number of candidate answers. Note that $\mathbf{y}_t^{dense}$ is a sigmoid output of the teacher model. As a result, $\hat{y}_{tn}$ contains a score of 1.0 for the ground-truth answer and soft scores ranging from 0 to 1 for the other candidates. Based on the combined labels $\hat{y}_{tn}$, we cast VisDial as a regression task that predicts the correctness of each candidate answer individually. The predicted score vector for $N$ candidates is computed as: 
\begin{align}
    \mathbf{s_t} &= \sigma\big(\mathbf{M}^a_t \mathbf{h}^\top_t\big)
\end{align}
where $\mathbf{M}^a_t \in \mathbb{R}^{N \times d_h}$ (in \sect~\hyperref[sec:input]{3.1}) and $\mathbf{h}_t \in \mathbb{R}^{1 \times d_h}$ are feature vectors for candidate answers and the hidden node feature for current round from SGL, respectively. $\sigma$ denotes a sigmoid function. Finally, we design a loss function for KT as:  
\begin{equation}
\mathcal{L}_{kt}=-\sum_{t=1}^{T} \sum_{n=1}^{N} \hat{y}_{tn} \ln \left(s_{tn}\right) -\left(1-\hat{y}_{tn}\right) \ln \left(1-s_{tn}\right)
\end{equation}
which is similar to a binary cross-entropy loss except that we use a \emph{soft} target score $\hat{y}_{tn}$. $\mathcal{L}_{kt}$ and the sigmoid activation function allow optimization for multiple correct answers. We believe KT is an efficient approach to distill the prior knowledge of dense labels from the teacher model for the training split, rather than directly fine-tuning the model on those dense labels only for validation split.


\section{Experiments}
\label{sec:experiments}
\subsection{Experimental Setup}
\noindent\textbf{Dataset.} We benchmark our proposed model on the VisDial v1.0 dataset \cite{das2017visual}. The VisDial v1.0 dataset contains 1.2M, 20k, and 44k question-answer pairs as train, validation, and test splits, respectively. The 123,287 images from COCO \cite{lin2014microsoft}, 2,064, and 8k images from Flickr are used to collect the dialog data for each split, respectively. A list of $N=$100 answer candidates accompanies each question-answer pair.

\noindent\textbf{Evaluation.} We follow the standard protocol \cite{das2017visual} for evaluating visual dialog models: mean reciprocal rank (MRR), recall@k (R@k), mean rank (Mean), and normalized discounted cumulative gain (NDCG). The first three measure the performance of retrieving the single ground-truth answer, while NDCG considers all relevant answers from the 100-answers list by using the densely annotated scores. There is a growing consensus among recent works \cite{kim2020modality,murahari2019large} that MRR and NDCG are regarded as the primary metrics and a balance of the two is important. For this reason, we additionally report the average of MRR and NDCG as \emph{overall} performance. The overall performance is also used as a selection criterion of VisDial challenge winner.

\subsection{Quantitative Analysis}
\noindent\textbf{Compared Methods.} We compare our methods with the state-of-the-art approaches on VisDial v1.0 dataset, including GNN \cite{zheng2019reasoning}, CorefNMN \cite{kottur2018visual}, RvA \cite{niu2018recursive}, Synergistic \cite{guo2019image}, ReDAN \cite{gan2019multi}, DAN \cite{kang2019dual}, HACAN \cite{Yang_2019_ICCV}, FGA \cite{schwartz2019factor}, MCA \cite{agarwal2020history}, P1+P2 \cite{qi2020two}, VisDial-BERT \cite{murahari2019large}, VD-BERT \cite{wang2020vd}.

\noindent\textbf{Comparison with State-of-the-art.} We evaluate our proposed methods with three different settings: (1) single model that utilizes the one-hot encoded labels (\ie, SGL), (2) single model with dense labels (\ie, SGL+KT), and (3) ensemble model with dense labels (\ie, $5\times$(SGL+KT)). As shown in \tbl~\ref{tab:t1}, (2) and (3) outperform the existing models on overall performance by 4.68\% (65.31 vs. 60.63) and 2.71\% (66.03 vs. 63.32), respectively. The results indicate that our methods show higher and more balanced performance than all other methods on NDCG and MRR. The single model also shows competitive performance compared with VD-BERT that utilizes BERT \cite{devlin2018bert} as a backbone. We observe that the use of dense labels yields huge improvements on NDCG and counter-effect on other metrics. Specifically, VD-BERT shows nearly 14\% improvements on NDCG with dense labels (59.96 $\rightarrow$ 74.54) while dramatically dropping MRR  (65.44 $\rightarrow$ 46.72). However, KT still boosts NDCG (61.97 $\rightarrow$ 72.60), yet notably with limited MRR drop (62.28 $\rightarrow$ 58.01). We conjecture that optimizing the loss on the combined labels (see \sect~\ref{sec:knowledge_transfer}) mitigates the counter-effect.
\begin{table}
\centering
\resizebox{\columnwidth}{!}{
\begin{tabular}{lccccccc}
\hline
\toprule
Model & Overall$\uparrow$ & NDCG$\uparrow$ & MRR$\uparrow$ & R@1$\uparrow$ & R@5$\uparrow$ & R@10$\uparrow$ & Mean$\downarrow$ \\
\midrule
GNN & 57.10 & 52.82 & 61.37 & 47.33 & 77.98 & 87.83 & 4.57 \\
CorefNMN & 58.10 & 54.70 & 61.50 & 47.55 & 78.10 & 88.80 & 4.40\\
RvA & 59.31 & 55.59 & 63.03 & 49.03 & 80.40 & 89.83 & 4.18\\ 
Synergistic & 59.76 & 57.32 & 62.20 & 47.90 & 80.43 & 89.95 & 4.17\\
Synergistic$\ddagger$ & 60.65 & 57.88 & 63.42 & 49.30 & 80.77 & 90.68 & 3.97 \\
ReDAN & 57.50 & 61.86 & 53.13 & 41.38 & 66.07 & 74.50 & 8.91 \\
ReDAN+$\ddagger$ & 59.10 & 64.47 & 53.73 & 42.45 & 64.68 & 75.68 & 6.63 \\
DAN & 60.40 & 57.59 & 63.20 & 49.63 & 79.75 & 89.35 & 4.30\\
DAN$\ddagger$ & 62.14 & 59.36 & 64.92 & 51.28 & 81.60 & 90.88 & 3.92 \\
HACAN & 60.70 & 57.17 & 64.22 & 50.88 & 80.63 & 89.45 & 4.20\\
FGA & 57.90 & 52.10 & 63.70 & 49.58 & 80.97 & 88.55 & 4.51\\
FGA$\ddagger$ & 60.90 & 54.50 & \textbf{67.30} & \textbf{53.40} & \textbf{85.28} & \textbf{92.70} & \textbf{3.54} \\
MCA$\dagger$ & 55.08 & 72.47 & 37.68 & 20.67 & 56.67 & 72.12 & 8.89 \\
P1+P2$\dagger$ & 60.09 & 71.60 & 48.58 & 35.98 & 62.08 & 77.23 & 7.48 \\
P1+P2$\dagger\ddagger$ & 63.32 & 74.02 & 52.62 & 40.03 & 68.85 & 79.15 & 6.76 \\
VisDial-BERT$\dagger$ & 62.60 & 74.47 & 50.74 & 37.95 & 64.13 & 80.00 & 6.28 \\
VD-BERT & 62.70 & 59.96 & 65.44 & 51.63 & 82.23 & 90.68 & 3.90 \\
VD-BERT$\dagger$ & 60.63 & 74.54 & 46.72 & 33.15 & 61.58 & 77.15 & 7.18 \\
VD-BERT$\dagger\ddagger$ & 63.26 & \textbf{75.35} & 51.17 & 38.90 & 62.82 & 77.98 & 6.69 \\
\midrule
SGL & 62.13 & 61.97 & 62.28 & 48.15 & 79.65 & 89.10 & 4.34  \\
SGL+KT$\dagger$ & 65.31 & 72.60 & 58.01 & 46.20 & 71.01 & 83.20 & 5.85  \\
SGL+KT${\dagger\ddagger}$ & \textbf{66.03} & 73.70 & 58.36 & 46.63 & 71.28 & 84.15 & 5.57 \\
\bottomrule
\hline
\end{tabular}
}
\caption{Test-std performance of the discriminative model on the VisDial v1.0 dataset. $\uparrow$ indicates higher is better. $\downarrow$ indicates lower is better. $\dagger$ denotes the use of dense labels. $\ddagger$ denotes ensemble model.}
\label{tab:t1}
\end{table}\\

\begin{table}
\centering
\resizebox{0.9\columnwidth}{!}{
\begin{tabular}{lccc}
\hline
\toprule
Model & Overall & NDCG & MRR \\
\midrule
Edgeless & 60.75 & 61.96 & 59.54 \\
Dense & 61.05 & 58.85 & 63.25 \\
Sparse-hard & 61.44 & 59.71 & 63.16 \\
P1+P2$\dagger$ (teacher model) & 61.65 & 73.42 & 49.88 \\
\midrule
SGL w/o RPN & 61.56 & 61.25 & 61.86 \\
SGL w/o SS & 61.66 & 62.46 & 60.85 \\
SGL w/o MR & 62.11 & 62.42 & 61.79 \\
SGL & 63.38 & 63.41 & \textbf{63.34} \\
SGL+KT$\dagger$ & \textbf{66.82} & \textbf{74.54} & 59.10 \\
\bottomrule
\hline
\end{tabular}
}
\caption{Comparison with the baseline models on the VisDial v1.0 validation split. MR, SS, and RPN denote the use of multi-step reasoning, structural supervision, and region proposal network, respectively. $\dagger$ denotes the use of dense labels.}
\label{tab:t2}
\end{table}

\begin{figure}
\begin{floatrow}
\ffigbox{%
  \centering
  \includegraphics[width=0.54\textwidth]{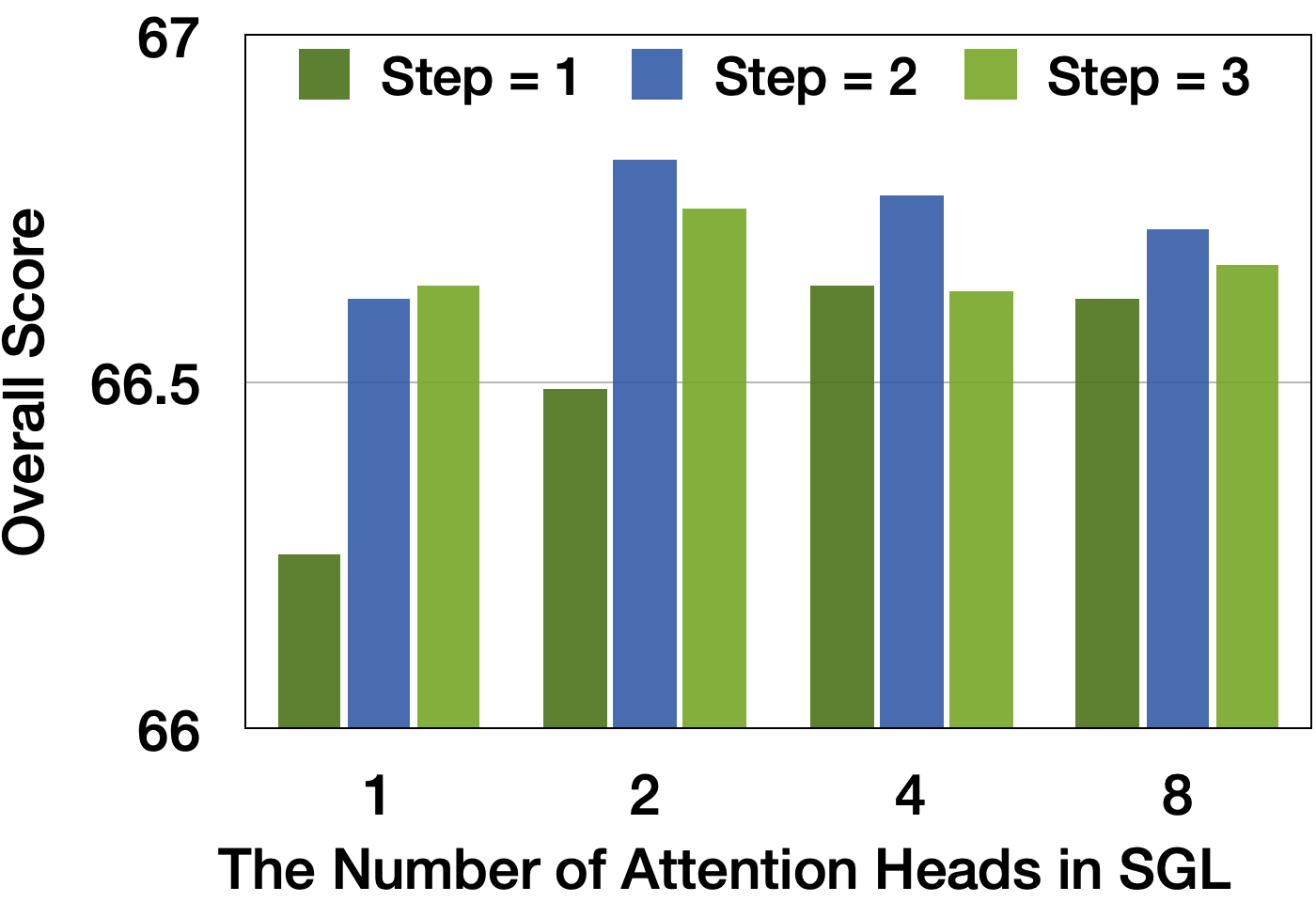}%
}{%
  \caption{Ablation study on VisDial v1.0 val split.}%
  \label{fig:ablation}
}
\capbtabbox{%

\resizebox{0.43\textwidth}{!}{
\begin{tabular}{lc}
\hline
\toprule
Model & F1-Score\\
\midrule
Edgeless & 0.0 \\
Dense & 0.246 \\
Sparse-hard & 0.279 \\ 
\midrule
SGL & 0.714 \\ 
SGL+KT & \textbf{0.748} \\ 
\bottomrule
\hline
\end{tabular}
}
}{%
  \caption{Graph inference on VisDial v1.0 val split.}%
  \label{tab:t3}
}
\end{floatrow}
\end{figure}

\begin{figure*}[h!]
\centering
\includegraphics[width=0.95\textwidth]{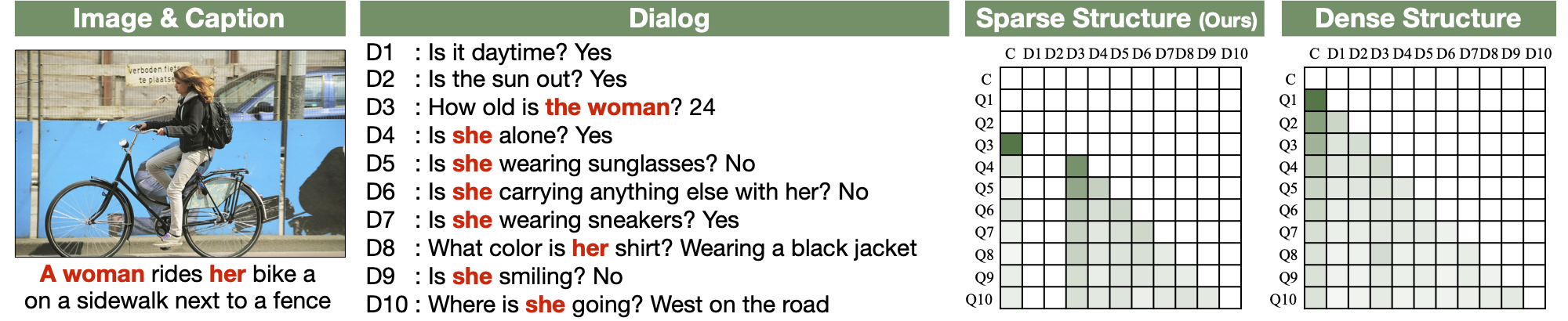}
\caption{A visualization of the inferred semantic structures from the validation set. From the left, the given image and caption, the dialog history, and the structures of ours and baseline. The darker the color, the higher the score.}
\label{fig:visualization1}
\end{figure*}
\begin{figure*}[h!]
\centering
\includegraphics[width=0.95\textwidth]{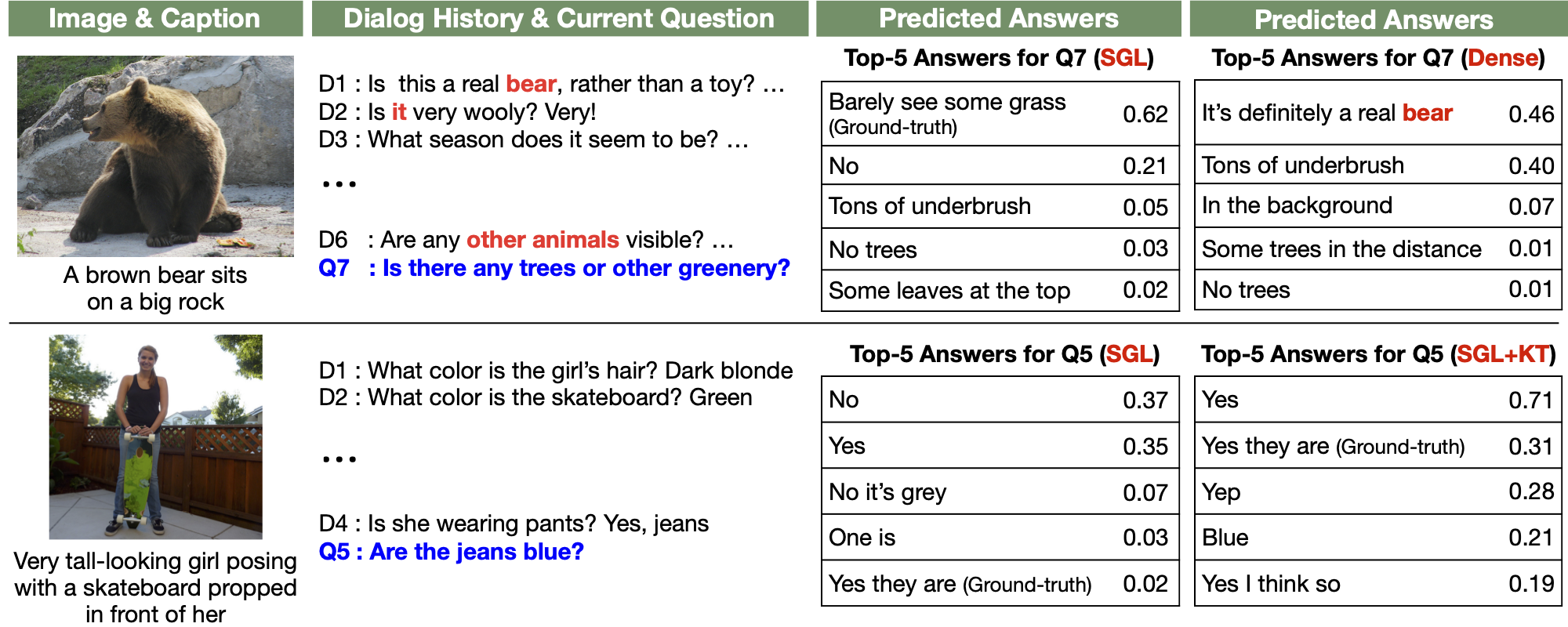}
\caption{A visualization of the top five predicted answers from SGL+KT, SGL, and Dense. Note that SGL+KT utilizes the sigmoid activation function to compute the answer scores while the others use the softmax function.}
\label{fig:visualization2}
\end{figure*}

\noindent\textbf{Comparison with Baselines.} We compare our methods to the baseline models in \tbl~\ref{tab:t2}. First, we define three models as baselines for SGL: Edgeless, Dense, and Sparse-hard. The Dense model utilizes a soft-attention mechanism, which yields the fully-connected graph. Contrary to the Dense model, the Sparse-hard model picks exactly one edge weights for each node by applying the Gumbel-Softmax to all nodes in the graph. Note that the structural supervision is provided in the Sparse-hard model. Finally, the Edgeless model yields a graph consisting only of isolated nodes. This indicates that the Edgeless model does not utilize the dialog history at all. As shown in \tbl~\ref{tab:t2}, SGL achieves better performance than the baseline models on all metrics. Furthermore, we report the performance of ablative models: SGL w/o RPN, SGL w/o SS, and SGL w/o MR. SGL w/o RPN employs ImageNet pre-trained with VGG-16 model \cite{simonyan2014very}, and uses the spatial grids of \emph{pool5} feature map as visual features. SGL w/o SS is the model that does not use the structural supervision (\ie, $L_{sgl}$). SGL w/o MR denotes the model that uses single-step reasoning in the sparse graph learning module. We identify that all three components (\ie, RPN, SS, and MR) in SGL play a crucial role in boosting the performance. Next, comparing SGL with SGL+KT, we observe that KT significantly improves NDCG score from 63.41 to 74.54. It demonstrates that the knowledge of the teacher model -- which helps to find multiple correct or relevant answers -- is successfully transferred to SGL. In \tbl~\ref{tab:t2}, SGL+KT even surpasses the NDCG score of the teacher model, P1+P2, by 1.12\%. From this observation, we conjecture that SGL enriches the distilled knowledge from the teacher model, which results in better performance than the teacher model. Although boosting NDCG results in decreasing MRR score due to their trade-off relationship \cite{murahari2019large,kim2020modality}, the MRR drop of KT is considerably smaller than other methods. 

\noindent\textbf{Reasoning Steps \& Attention Heads.} Based on SGL+KT model, we perform ablation experiments with different number of reasoning steps (1, 2, and 3) in the sparse graph learning module and attention heads (1, 2, 4, and 8) in the node embedding module. As shown in \fig~\ref{fig:ablation}, the model with two-step reasoning with two attention heads performs the best among all models in the experiments, recording 66.82 on overall performance.    

\noindent\textbf{Is SGL inferring the right graph?} We investigate this question by measuring the agreement between the binary edges $\mathbf{A}^b$ inferred from our model and the structural supervision $\mathbf{C}$, assuming that $\mathbf{C}$ is the ground-truth graph. We use F1-score as an evaluation metric. Then, we employ Edgeless, Dense, and Sparse-hard as baselines. Note that the Dense model itself is not compatible with the evaluation metric since it does not predict the binary edges. To make it compatible, we create the binary edges by replacing the top edge weight for each node with one. The rest are replaced with zero. In \tbl~\ref{tab:t3}, SGL and SGL+KT show significantly better F1-scores than the baselines. It might indicate that SGL infers more reliable semantic structures. Furthermore, comparing SGL with SGL+KT in \tbl~\ref{tab:t3}, we observe that KT improves the performance of graph inference. It indicates that KT contributes to an accurate inference of sparse graphs. 

\subsection{Qualitative Results}
In \fig~\ref{fig:visualization1}, we visualize the images, the corresponding dialogs in the validation split, and the inferred adjacency matrices as well as the ones from the Dense model as a counter. Compared to the dense structure in the baseline, the proposed SGL indeed learns the innate sparse structures, and the question nodes receive the information from the other nodes in a selective fashion. For instance, the questions from Q3 to Q10 have non-zero binary edges to all previous contexts except D1 and D2, which do not contain relevant information about `the woman'. On the contrary, Q1 and Q2 are not connected to any other nodes, because they can be answered solely without additional context. We visualize additional examples regarding the graph inference in the supplementary materials. Next, to demonstrate the advantages of SGL and KT, we visualize the top five predicted answers for each question from the Dense model, SGL, and SGL+KT in \fig~\ref{fig:visualization2}. In the first example, SGL retrieves the ground-truth answer by not using the dialog history, while the Dense model provides the wrong answer -- containing the word \emph{bear} -- to the top. We conjecture that relying on the dialog history -- even when the history is not required -- leads to this phenomenon. In the next example, the answers predicted by SGL+KT are semantically exchangeable with each other, whereas the answers from SGL are not. It shows that the teacher model’s knowledge enforces the ability to find multiple correct answers and resultant consistency of answer prediction.
\section{Conclusions}
We propose SGL and KT to remedy the shortcomings of previous work: soft-attention and one-hot labels. Experimental results illustrate the effectiveness of our approach. SGL with KT achieves the new state-of-the-art performance on the VisDial v1.0 dataset. We believe that the idea of selectively paying attention to desired information is widely applicable to various research fields, and KT can be generally adopted to improve answer prediction.

\paragraph{Acknowledgement}
The authors would like to thank Woo-Suk Choi and Björn Bebensee for helpful comments and editing. This work was supported in part by SK Telecom when Gi-Cheon Kang, Hwaran Lee, and Jin-Hwa Kim worked at SK Telecom. The Korean government (2015-0-00310-SW.StarLab, 2017-0-01772-VTT, 2018-0-00622-RMI, 2019-0-01371-BabyMind) partly supports this work as well.

\bibliography{anthology,custom}
\bibliographystyle{acl_natbib}
\clearpage
\appendix
{\Large \noindent\textbf{Appendix Overview}} \\

\noindent The supplementary materials are organized as:
\begin{itemize}
  \item \sect~\hyperref[sec:ane]{A} shows a detailed architecture of the node embedding module. 
  \item \sect~\hyperref[sec:agm]{B} presents our experiments with a generative model.
  \item \sect~\hyperref[sec:aid]{C} presents implementation details.
  \item \sect~\hyperref[sec:aqe]{D} shows qualitative examples from SGL.
\end{itemize} 

\section{Node Embedding Module}
\label{sec:ane}
\begin{figure}
    \centering
    \includegraphics[width=0.95\textwidth]{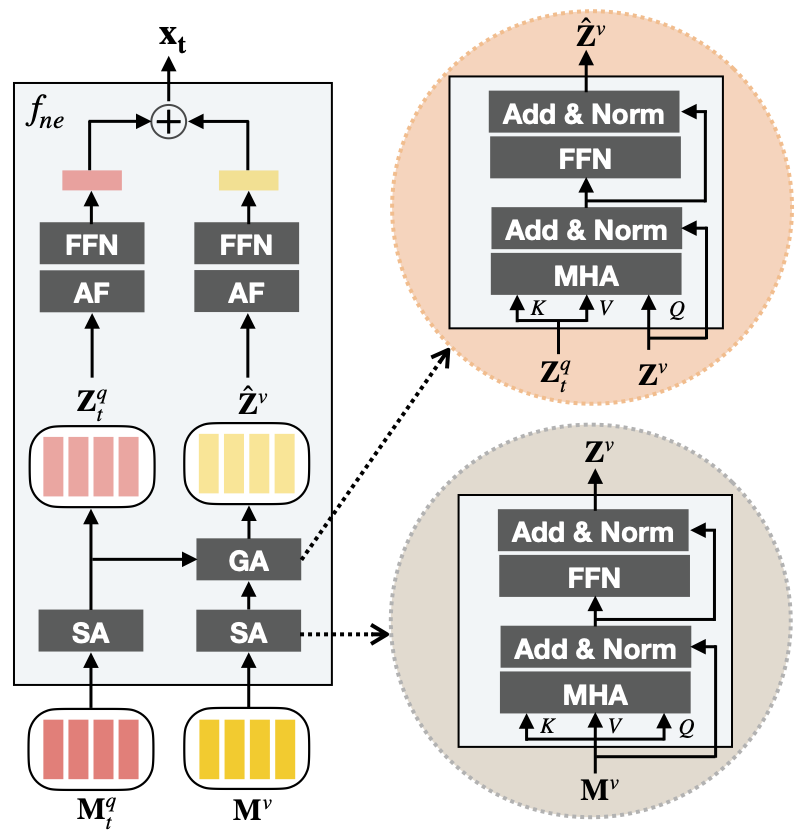}
    \caption{A detailed architecture of the node embedding module. SA, GA, AF, MHA, and FFN denote self-attention, guided-attention, attention flat, multi-head attention, and feed-forward networks, respectively.}
    \label{fig:ne}
\end{figure}

\noindent\textbf{Subcomponents.} A detailed architecture of the node embedding module is presented in Figure~\ref{fig:ne}. The module consists of three subcomponents: self-attention (\ie, SA), guided-attention (\ie, GA), and attention flat (\ie, AF). First, SA and GA are based on the multi-head attention mechanism (\ie, MHA) \cite{vaswani2017attention}. MHA computes $h$ parallel attention heads and aggregates them with a linear matrix. Each head corresponds to the output of the scaled dot-product attention. It is formulated as: 
\begin{gather}
    \mathrm{ A(} {\mathbf{Q}, \mathbf{K}, \mathbf{V}}  \mathrm{)} = \mathrm{softmax(} { \frac{{\mathbf{QK}}^\top}{\sqrt{{d}}} \mathrm{)}}{\mathbf{V}} \\
    \mathrm{ MHA(} {\mathbf{Q}, \mathbf{K}, \mathbf{V}}  \mathrm{)} = [head_1, ... , head_h]\mathbf{W}^o \\
    head_n  = \mathrm{A (} {\mathbf{Q}\mathbf{W}^Q_{n}}, {\mathbf{K}\mathbf{W}^K_n}, {\mathbf{V}\mathbf{W}^V_n} \mathrm{)}
\end{gather}
where $\mathbf{W}^Q_{n}, \mathbf{W}^K_{n}, \mathbf{W}^V_{n}$ are the projection matrices for the $n$-th head. $\mathbf{W}^O$ is the linear matrix. Then, the residual connection \cite{he2016deep}, layer normalization \cite{ba2016layer}, and the two-layer feed-forward networks (\ie, FFN) are applied in SA and GA (see Figure~\ref{fig:ne}). The inputs of SA are from the same features, while GA takes two groups of input features -- the query and the key-value pairs. Next, AF performs an attentional reduction to flatten the inputs to the vector representation. Given the input matrix $\textbf{X} = [\textbf{x}_1, \cdots, \textbf{x}_m]\in$ $\mathbb{R}^{m \times d_h}$, AF yields the vector $\tilde{\textbf{x}} \in$ $\mathbb{R}^{1 \times d_h}$ as follows: 
\begin{align}
    \mathrm{AF(}\textbf{X}\mathrm{)} = \tilde{\textbf{x}} = \sum_{i=1}^m \alpha_i \textbf{x}_i \\
    \alpha = \mathrm{softmax(MLP(} {\textbf{X}} \mathrm{))}
\end{align}
where MLP projects \textbf{X} to $m$-dimensional vector. $\alpha = [\alpha_1, \cdots, \alpha_m] \in$ $\mathbb{R}^{m}$ are the attention weights. 
\\

\noindent\textbf{Overview.} First, the object-level visual features $ \mathbf{M}^v \in$ $\mathbb{R}^{K \times d_h}$ and the question features $\mathbf{M}^q_t \in$ $\mathbb{R}^{L \times d_h}$ are given to SA, yielding $\textbf{Z}^v \in$ $\mathbb{R}^{K \times d_h}$ and $\textbf{Z}^q_t \in$ $\mathbb{R}^{L \times d_h}$, respectively. Then, GA takes $\textbf{Z}^v$ and $\textbf{Z}^q_t$ as inputs and computes the pair-wise relationship between the visual features and the linguistic features. $\hat{\mathbf{Z}}^v \in$ $\mathbb{R}^{K \times d_h}$ is obtained from GA. Finally, $\hat{\mathbf{Z}}^v$ and $\textbf{Z}^q_t$ are passed through to AF($\cdot$) and the two-layer feed-forward networks, resulting in $\textbf{z}^v \in$ $\mathbb{R}^{1 \times d_h}$ and $\textbf{z}^q_t \in$ $\mathbb{R}^{1 \times d_h}$, respectively. Consequently, the visual-linguistic representation $\textbf{x}_t \in$ $\mathbb{R}^{1 \times d_h}$ is obtained by adding $\textbf{z}^v$ and $\textbf{z}^q_t$. From this pipeline, the node embedding module $f_{ne}$ embeds the high-level abstraction of the visual and linguistic inputs in a joint fashion. Note that the module also embeds each round of the dialog history in the same way as the question features. 
\section{Generative Model}
\label{sec:agm}
\noindent\textbf{Overview.} The authors of \cite{das2017visual} have also proposed a generative model which is trained for generating an answer without access to the answer candidates. Specifically, the generative model aims to generate the ground-truth answer's word sequence auto-regressively via a LSTM:
\begin{align}
\label{eq:generativemodel}
\begin{split}
\mathcal{L}_{gen} &= -\sum_{t=1}^{T} \mathrm{log}\;p(a^{gt}_t|\mathbf{h}_t) \\
&= -\sum_{t=1}^{T} \sum_{l=1}^{L} \mathrm{log}\;p({w}_{l}|{w}_{<l},\mathbf{h}_{t})
\end{split}
\end{align}
where $\mathbf{h}_t$ is the hidden node feature for the current round from SGL and $a^{gt}_t$ denotes the ground-truth answer consisting of $L$ words $({w}_{1}, ..., {w}_{L})$. $T$ is the number of rounds for each dialog. We initialize the hidden states of the LSTM with $\mathbf{h}_t$. Then, the generative model is optimized by minimizing negative log-likelihood of the ground-truth answer. In inference time, following \citet{das2017visual}, we utilize the log-likelihood scores to determine the rank of candidate answers for the process of evaluation.
\\

\noindent\textbf{Generative Model with Knowledge Transfer.} We further apply the Knowledge Transfer (KT) technique to the generative model. Based on the combined labels $\hat{\mathbf{y}}_t$, which were discussed in \sect~\ref{sec:knowledge_transfer}, we extract the top-$I$ answer candidates for the given question and use them to train the model. Formally,  
\begin{align}
\begin{split}
    \mathcal{L}_{gen, \, kt} &= -\sum_{t=1}^{T} \hat{\mathbf{y}}_t \; \mathrm{log}\;p(\hat{\mathcal{A}}_t|\mathbf{h}_t) \\
    &= -\sum_{t=1}^{T} \sum_{i=1}^{I} \hat{y}_{ti}\; \mathrm{log}\;p(a_{t}^i|\mathbf{h}_t) \\
    &= -\sum_{t=1}^{T} \sum_{i=1}^{I} \hat{y}_{ti} \sum_{l=1}^{L} \mathrm{log}\;p({w}_{i,l}|{w}_{i,<l},\mathbf{h}_{t})
\end{split} 
\end{align}
where $\hat{\mathcal{A}}_t = \left\{a_t^i \right\}_{i=1}^{I}$ is a set of selected candidate answers and $a_t^i$ consists of $L$ words $({w}_{i,1}, ..., {w}_{i,L})$. $I$ implies the number of candidate answers that the generative model can access. Accordingly, $I=1$ is equivalent to the standard generative model described in Eq.~\ref{eq:generativemodel} since the ground-truth answer contains the highest score (\ie, 1.0). Note that $\mathcal{L}_{gen, \, kt}$ computes a \emph{weighted} negative log-likelihood loss because each selected candidate answer $a_t^i$ has a different confidence score. 
\begin{table}
\centering
\resizebox{\columnwidth}{!}{
\begin{tabular}{lccccccc}
\hline
\toprule
Model & Overall & NDCG & MRR \\
\midrule
MN \cite{das2017visual} & 52.41 & 56.99 & 47.83 \\ 
HCIAE \cite{lu2017best} & 54.39 & 59.70  & 49.07\\
ReDAN \cite{gan2019multi} & 55.25 & 60.47 & {\bf 50.02}
\\
\midrule
SGL & 55.30 & 61.42 & 49.17 \\
SGL+KT ($I=2$) & 55.42 & 63.80 & 47.03 \\
SGL+KT ($I=3$) & \textbf{56.21} & \textbf{65.74} & 46.67 \\
\bottomrule
\hline
\end{tabular}
}
\caption{VisDial v1.0 validation performance of the generative models.}
\label{tab:t4}
\end{table}
\\

\noindent\textbf{Experimental Results.} We report the performance of the generative model on the VisDial v1.0 validation split. As shown in \tbl~\ref{tab:t4}, SGL shows slightly better performance than ReDAN \cite{gan2019multi} on overall performance. Furthermore, we find that only a small subset of the knowledge of the teacher model is also effective for this generative approach. As observed in the discriminative model in \sect~5, the use of teacher knowledge also leads to huge NDCG improvements and the counter-effect on other metrics.   

\section{Implementation Details}
\label{sec:aid}
We use pre-trained Glove \cite{pennington2014glove} to embed all the language inputs. The maximum sequence length of the questions, answers, and captions is 20, 20, and 40, respectively. Based on this maximum length, each language input is padded or truncated. We use $K=10 \sim 100$ object-level visual features for reflecting the complexity of each image. The dimension of each feature is $d_v=2048$ and the number of attention heads in multi-head attention is $h=2$. The dimension of $d_h$ is 512. The total number of rounds for each dialog $T$ is 10 and the number of candidate answers $N$ is 100. The softmax temperature for computing the binary edges $\tau$ is 0.5. We employ the Adam optimizer \cite{kingma2014adam} with initial learning rate $1 \times 10^{-4}$. The learning rate is warmed up to $4 \times 10^{-4}$ until epoch 4 and is halved every three epochs from 12 to 24 epochs.

\section{Qualitative Examples}
\label{sec:aqe}
\begin{figure*}[h!]
\centering
\includegraphics[width=\textwidth]{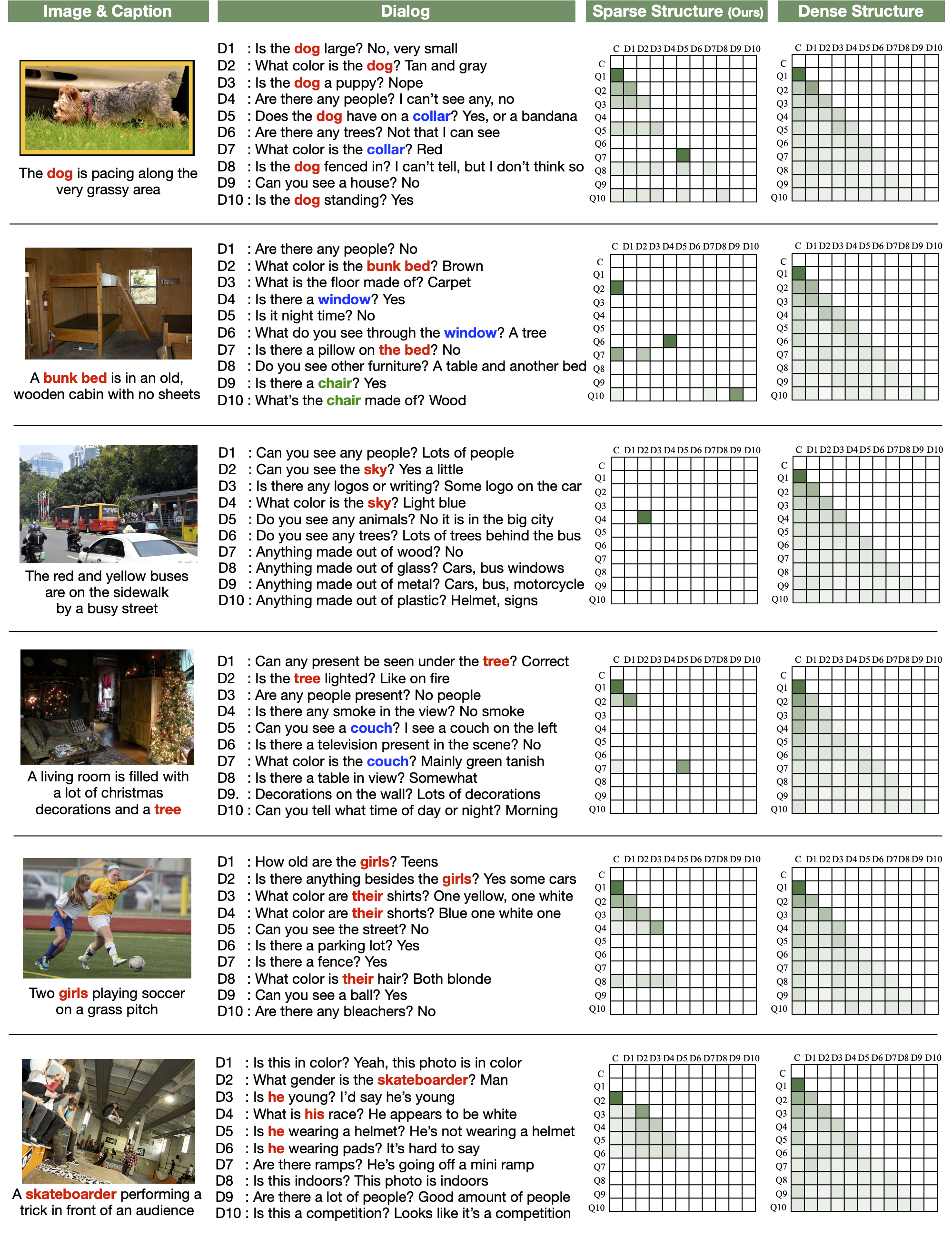}
\caption{The additional examples of the inferred semantic structures from the validation split. From the left, the given image and caption, the dialog history, and the structures of ours and the baseline. The darker the color, the higher the score.}
\label{fig:visualization3}
\end{figure*}

We visualize the inferred graph structures from our proposed model and the ones from the Dense model as a comparison. As shown in Figure~\ref{fig:visualization3}, our proposed model indeed captures semantic structures among a series of utterances by selectively attending to the dialog history. On the other hand, the Dense model yields fully-connected graphs due to two constraints of the softmax function: (1) the softmax function always assigns non-zero values to all edge weights, and (2) the sum of the edge weights for each node should be one. However, SGL can assign zero values to all edge weights if needed (\eg, Q4 in the first example of Fig.~\ref{fig:visualization3}). We believe this ability is crucial to prevent the visual dialog model from overly relying on the dialog history. 

\end{document}